\documentclass[sigconf,nonacm]{acmart}

\setcopyright{none}
\renewcommand\footnotetextcopyrightpermission[1]{}
\acmConference[KDD~'27]{Proceedings of the 33rd ACM SIGKDD Conference on Knowledge Discovery and Data Mining}{August 1--5, 2027}{San Jose, CA, USA}
\acmYear{2027}
\copyrightyear{2027}
\microtypesetup{expansion=false,protrusion=false}
\usepackage{booktabs}
\usepackage{amsmath}

\usepackage{amssymb}
\usepackage{xcolor}
\usepackage{pifont}
\usepackage{graphicx}
\usepackage{algorithm}
\usepackage{algorithmic}
\usepackage{tikz}
\usetikzlibrary{arrows.meta,positioning,fit,backgrounds}
\newcommand{\cmark}{\textcolor{green!55!black}{\ding{51}}}
\newcommand{\xmark}{\textcolor{red!70!black}{\ding{55}}}
\newcommand{\suite}[1]{\textsf{#1}}
\newcommand{\code}[1]{\texttt{#1}}
\newcommand{\gapnote}{Gaps are computed from unrounded values; displayed cells are rounded independently, so a cell-wise subtraction can differ by up to $0.01$.}
\newcommand{\tgapnote}{Gap SEs combine the two arms' SEs in quadrature; both arms score the same seeded episode stream, so under positively correlated episode difficulty this is conservative. The four $T$ columns roll the same evaluation episode population out to longer horizons (fixed evaluation seed), isolating horizon length rather than providing independent replications.}
\newcommand{\sd}[1]{\,{\scriptscriptstyle\pm#1}}

\begin{document}

\title{DoTime: A Synthetic Benchmark Generator for Interventional and Counterfactual Time Series}

\author{Dennis Thumm}
\affiliation{%
  \institution{National University of Singapore}
  \city{Singapore}\country{Singapore}}
\email{dennis.thumm@u.nus.edu}

\author{Billy Tim Anthony}
\affiliation{%
  \institution{National University of Singapore}
  \city{Singapore}\country{Singapore}}
  \email{tim.anthony@u.nus.edu}

\author{Ying Chen}
\affiliation{%
  \institution{National University of Singapore}
  \city{Singapore}\country{Singapore}}
  \email{matcheny@nus.edu.sg}

\begin{abstract}
Most benchmarks for causal inference over time series are observational, small, or domain-specific, leaving interventional and counterfactual estimation under-served exactly where it matters most, such as in healthcare, policy evaluation, and climate science. 
We introduce \textbf{DoTime}, an open, scalable, and theoretically grounded generator of multivariate temporal structural causal models (TSCMs) with interventions, released as the \code{dotime} PyPI package together with four frozen evaluation suites. 
Beyond existing work, it adds capabilities absent from prior generators: continuous-time intervention \emph{windows}, counterfactual sampling modes with a positivity guard, regime-switching SCMs as a strict generalization of interrupted time series, non-stationary dynamics by construction with switching SCM parameters, and deterministic ramp and sinusoidal intervention profiles that place trends and structural breaks \emph{inside} the evaluation window. Moreover, it demonstrates the suitability of the generator as a prior for a causal foundation model reference implementation.
The released suites span a training-scale snapshot of $100{,}000$ trajectories and eight named identification structures, each with exact ground truth: paired interventional trajectories from the same SCM throughout, and shared-noise counterfactuals in the continuous-time suite. We ship reference baseline implementations with an evaluation harness, and pose a falsifiable claim:  interventional training buys a measurable direction-accuracy advantage over an observational model of identical capacity. It is tested across three training seeds per arm. Under structure-matched evaluation on held-out episodes, the interventional prior-fitted network's (PFN) gap is positive in every structure, trajectory length, and seed tested.
\end{abstract}

\keywords{causal inference, time series, interventions, counterfactuals,
benchmarks, structural causal models, foundation models}

\ccsdesc[500]{Computing methodologies~Causal reasoning and diagnostics}
\ccsdesc[500]{Computing methodologies~Machine learning}
\ccsdesc[300]{Mathematics of computing~Time series analysis}
\ccsdesc[300]{Information systems~Test collections}

\maketitle

\section{Introduction}\label{sec:intro}

Counterfactual reasoning over time is the bottleneck in many high-stakes domains.
Estimating the effect of a treatment policy on a patient's trajectory, of a non-pharmaceutical intervention on an epidemic curve, or of an emissions change on a climate index all require answering $\mathrm{do}(\cdot)$ queries about how a \emph{multivariate} system would have evolved under intervention. 
Progress on such questions is increasingly driven by \emph{prior-fitted networks} (PFNs) and tabular/time
foundation models~\cite{muller2022tabpfn,tabpfn2023,robertson2025dopfn} that are trained entirely on synthetic data sampled from a structural prior. 
To facilitate this progress, \textbf{DoTime} provides two distinct resources doing two distinct jobs: the prior determines what a model can learn, while the benchmark determines what a reader can detect about it. Both are therefore decisive, and both are affected by the same shortcoming. The resources the community trains and evaluates on are overwhelmingly \emph{observational}, so a model trained to answer $p(Y \mid \mathrm{do}(A))$ is fitted to, and scored on, data that only ever shows $p(Y \mid A)$.

This skew shows up along three axes. (a)~Observational time-series generators such as Dream3~\cite{dream3} and NetSim~\cite{netsim2011} target structural discovery, not effect estimation. 
(b)~Domain-specific interventional datasets such as the Causal Chambers~\cite{causalchamber2024} are real but small and do not scale to foundation-model training. 
(c)~Static-tabular causal benchmarks such as ACIC~\cite{acic2019} and IHDP~\cite{ihdp2011} provide ground-truth effects but ignore time entirely. The recent \textsc{CausalDynamics} benchmark~\cite{herdeanu2025causaldynamics} scales to coupled ODE/SDE dynamics but evaluates \emph{structural discovery},  recovering a graph from observational dynamics, which is complementary to, and disjoint from, interventional effect estimation. \citet{poinsot2025position} document the absence of a widely agreed benchmark for time-series causal inference.

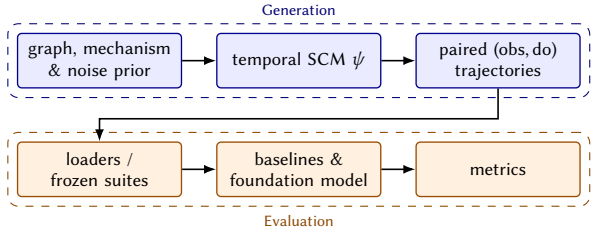
\begin{figure}[t]
\centering
\begin{tikzpicture}[
    font=\footnotesize\sffamily,
    node distance=3.2mm and 4.5mm,
    box/.style={draw, rounded corners=1.6pt, align=center, inner sep=2.4pt,
                minimum height=7.2mm, text width=20mm, line width=0.5pt},
    gen/.style={box, fill=blue!8, draw=blue!55!black},
    evl/.style={box, fill=orange!12, draw=orange!60!black},
    flow/.style={-{Latex[length=1.6mm]}, line width=0.6pt},
  ]
  \node[gen] (prior) {graph, mechanism \& noise prior};
  \node[gen, right=of prior] (scm) {temporal SCM $\psi$};
  \node[gen, right=of scm] (traj) {paired (obs,\,do) trajectories};
  \node[evl, below=7mm of prior] (suites) {loaders / frozen suites};
  \node[evl, right=of suites] (base) {baselines \& foundation model};
  \node[evl, right=of base] (metrics) {metrics};
  \draw[flow] (prior) -- (scm);
  \draw[flow] (scm) -- (traj);
  \draw[flow] (traj.south) |- ([yshift=3mm]suites.north) -- (suites.north);
  \draw[flow] (suites) -- (base);
  \draw[flow] (base) -- (metrics);
  \begin{scope}[on background layer]
    \node[draw=blue!45!black, dashed, rounded corners, inner sep=3.4pt,
          fit=(prior)(traj), label={[blue!45!black,font=\scriptsize\sffamily]above:Generation}] {};
    \node[draw=orange!55!black, dashed, rounded corners, inner sep=3.4pt,
          fit=(suites)(metrics), label={[orange!55!black,font=\scriptsize\sffamily]below:Evaluation}] {};
  \end{scope}
\end{tikzpicture}
\caption{The DoTime pipeline: a structural prior samples TSCMs, which are simulated observationally and under intervention to yield paired trajectories with exact ground-truth targets (noise-matched counterfactuals in the continuous-time generator); frozen suites and an evaluation harness close the loop.}
\label{fig:overview}
\end{figure}

\paragraph{Contribution as artifact.} We release the \code{dotime} PyPI package (generator, suite loaders, and evaluation harness), four frozen suites with Zenodo DOIs and Croissant metadata, a Hugging Face mirror, reference baselines, and a versioned public reference table with a scripted submission path (results are contributed via a GitHub pull request; \S\ref{sec:limitations}). Our contribution is the \emph{benchmark ecosystem}\footnote{Where \citet{thumm2026causal} introduced a single Do-Over-Time-PFN model instance, we generalize that sampling surface into an open generator and use the PFN here only as a \emph{reference implementation} that exercises the benchmark and validates its utility --- not as the paper's technical contribution. The convergence result (Appendix~\ref{app:theory}) is likewise restated to make this paper self-contained, not claimed as new theory.}: the generalized DoTime generator, the four evaluation suites, and the benchmarking protocol.
\paragraph{Contribution as benchmark scope.} We organize the design along four axes: the intervention type, structural family, counterfactual mode, and time grid (\S\ref{sec:scope}). Each is exercised by a released suite. 
\paragraph{Why principled.} The generator enforces acyclicity by construction and a positivity guard, with value clipping and divergence detection at simulation time. An optional hardening configuration additionally rescales lag matrices to a target spectral radius (used for PFN training, not applied to the released v1.0.0 suites). The underlying convergence result is restated in Appendix~\ref{app:theory}.
\paragraph{What our benchmark supports and what it does not.} DoTime measures interventional / counterfactual effect estimation on synthetic multivariate Temporal Structural Causal Models (TSCMs)~\cite{thumm2026interventional} with known ground truth, across named identification structures. It does \emph{not} validate real-world transportability or partial observability beyond the provided hidden-confounding case. These are explicit limitations (\S\ref{sec:limitations}), though it \emph{does} cover irregular observation grids (\S\ref{sec:scope}). Figure~\ref{fig:overview} summarizes the pipeline.

\section{Related Work and Positioning}\label{sec:related}

\paragraph{Static causal benchmarks.} ACIC~\cite{acic2019} and IHDP~\cite{ihdp2011} provide ground-truth effects but no temporal structure, and so cannot probe lagged confounding, intervention timing, or effect decay.

\paragraph{Time-series causal evaluation.} Dream3~\cite{dream3} and NetSim~\cite{netsim2011} are observational, the Causal Chambers~\cite{causalchamber2024} are real but small, and interventional Electronic Health Record (EHR) cohorts are proprietary. \textsc{CausalDynamics}~\cite{herdeanu2025causaldynamics} recovers graphs from observational dynamics over coupled ODE/SDE systems, and \textsc{CausalTime}~\cite{cheng2024causaltime} generates realistic observational series (air-quality, traffic, medical) for causal \emph{discovery}: both reconstruct graphs from observational data, whereas DoTime answers $\mathrm{do}(\cdot)$ queries over time\footnote{Despite the one-edit-distance name, \textsc{DoTime} $\neq$ \textsc{CausalTime}: the former applies the $\mathrm{do}$-operator over time, the latter produces realistic observational series with no interventions.}. Even where interventional temporal data appears in the discovery literature --- CAnDOIT~\cite{castri2024candoit} combines observational and interventional series to recover graphs --- the goal is structure identification, without counterfactual ground truth, frozen suites, or an effect-estimation protocol. Table~\ref{tab:comparison} summarizes the coverage gap.

\paragraph{Synthetic priors for foundation models.} Do-PFN~\cite{robertson2025dopfn} (static causal effects) and TempoPFN~\cite{tempopfn2025} (observational, univariate series) are the closest priors; DoTime is their natural \emph{temporal-interventional} generalization and extends the Do-PFN sampling surface for the instantaneous graph and mechanism priors. TSCM priors are meanwhile becoming mainstream in foundation-model pretraining --- TabPFN-3's discrete-time dynamic SCM~\cite{grinsztajn2026tabpfn3}, Google's TabFM pretrained on hundreds of millions of on-the-fly SCM datasets~\cite{tabfm2026}, Chronos-2's autoregressive rollouts of random temporal causal graphs~\cite{ansari2025chronos2} --- but none adds the interventional/counterfactual side DoTime supplies.

\begin{table}[t]
\caption{Where DoTime sits. \cmark/\xmark{} denote whether a benchmark
addresses each property: temporal, interventional, counterfactual, named
identification structures, configurable scale, and a citable DOI. DoTime is the only
entry covering interventional \emph{and} counterfactual estimation with named
structures at configurable scale.}
\label{tab:comparison}
\small
\setlength{\tabcolsep}{3.2pt}
\begin{tabular}{lcccccc}
\toprule
Benchmark & Temp. & Interv. & Counterf. & Struct. & Scale & DOI \\
\midrule
ACIC / IHDP            & \xmark & \cmark & \cmark & \xmark & \xmark & \cmark \\
Dream3 / NetSim        & \cmark & \xmark & \xmark & \xmark & \xmark & \cmark \\
Causal Chambers        & \cmark & \cmark & \xmark & \xmark & \xmark & \cmark \\
CausalDynamics         & \cmark & \xmark & \xmark & \cmark & \cmark & \cmark \\
CausalTime             & \cmark & \xmark & \xmark & \xmark & \xmark & \cmark \\
\textbf{DoTime} & \cmark & \cmark & \cmark & \cmark & \cmark & \cmark \\
\bottomrule
\end{tabular}
\end{table}

\section{The DoTime Framework}\label{sec:framework}

\subsection{Temporal Structural Causal Model family}
A TSCM~\cite{thumm2026interventional} is $\psi = (\mathcal{G}, f, p_\varepsilon)$, with graph collection $\mathcal{G}=(G_0,G_{1:K})$, an instantaneous Directed Acyclic Graph (DAG) $G_0$ over $N$ variables, lagged adjacencies $G_1,\dots,G_K$, a per-node mechanism $f_i$, and a noise distribution $p_\varepsilon$. The forward recursion is
\begin{equation}\label{eq:recursion}
X^i_t = f_i\!\left( \mathrm{Pa}^0_i(X_t),\, \mathrm{Pa}^{1:K}_i(X_{t-1:t-K}) \right) + \varepsilon^i_t,
\end{equation}
where $\mathrm{Pa}^0_i$ are instantaneous parents and $\mathrm{Pa}^{k}_i$ are lag-$k$ parents. We assume acyclicity of $G_0$ and Markovianity of order $K$; the theory additionally assumes a stable spectral radius, which the optional hardening configuration enforces by rescaling (in the released suites, stability is instead handled by clipping plus divergence detection, \S\ref{sec:validation}). Interventions act on Eq.~\eqref{eq:recursion} by overriding $f_i$ for the targeted node over a set of time steps (\S\ref{sec:scope}). 
This yields two distinct kinds of ground truth, along Pearl's hierarchy~\cite{pearl2009causality}. 
An \emph{interventional} target (Level~2, $\mathbb{E}[Y\mid\mathrm{do}(A{=}v)]$) draws the intervened arm's exogenous noise afresh from the prior $P(\varepsilon)$: it is forward-looking, and the two arms diverge from the first step because the intervened arm follows a different noise realisation. 
A \emph{counterfactual} target (Level~3, $\mathbb{E}[Y_{v}\mid A{=}a, Y{=}y]$) instead conditions on the observed run: the intervened arm reuses the \emph{same} noise stream $\varepsilon$ that produced the observational trajectory --- equivalently, the posterior $P(\varepsilon\mid\text{observed run})$, which for a fully observed SCM is that one realisation --- so the arms are bit-identical until the intervention and differ \emph{only} through it. 
The released continuous-time suite provides Level~3 counterfactuals by construction; the discrete-time suites provide Level~2 interventional twins (\S\ref{sec:limitations}).

\subsection{Graph prior}
$G_0$ is Erd\H{o}s--R\'enyi~\cite{erdos1960evolution} with acyclicity guaranteed by construction: edges are sampled from a strictly upper-triangular Bernoulli mask over a random topological order, so no rejection sampling is needed. Lagged edges are Bernoulli with geometric decay $\gamma^k$ in the lag $k$. Edge probability is drawn from $\mathrm{Beta}(2,5)$; $N\!\sim\!\mathcal{U}\{3,\dots,10\}$ and $K\!\sim\!\mathcal{U}\{1,\dots,3\}$. A subset of nodes is marked hidden to induce unobserved confounding. Figure~\ref{fig:structures} shows all eight identification structures of \suite{dot-Identifiability-v1} \emph{unrolled over time}: each edge of the identification skeleton is instantiated either instantaneously (within slice $t$) or as a lagged edge $G_{1:K}$ connecting earlier slices, and the intervention acts on $A_t$ (red) while the query reads $Y_T$.

\subsection{Mechanism and noise priors}
Mechanisms are drawn uniformly from $\{\text{linear}, \tanh, \sin, \cos, |\cdot|, \allowbreak (\cdot)^2, \mathrm{ReLU}, \tanh\!\circ(\cdot)^2, \tanh\!\circ\mathrm{ReLU}\}$, a set chosen to span monotone, saturating, periodic, and even nonlinearities. Root and non-root noise scales are sampled from shifted exponentials. Numerical stability is enforced by value clipping, a divergence check that replaces a diverged trajectory with an all-zero episode, and a burn-in that discards transient dynamics. The resulting zeroed fraction per released suite is reported and discussed in \S\ref{sec:validation}.

\subsection{Sampling pipeline}
A type sampler draws $70\%$ \emph{diverse} (random DAGs over the full mechanism set), $15\%$ \emph{chain} (a single directed path, the hardest long-range-propagation case), and $15\%$ regime-switching SCMs; each is simulated forward with burn-in to a configurable length $T$.
Algorithm~\ref{alg:sample} (Appendix~\ref{app:algo}) gives the full pseudocode and Algorithm~\ref{alg:cf} the counterfactual reconstruction.

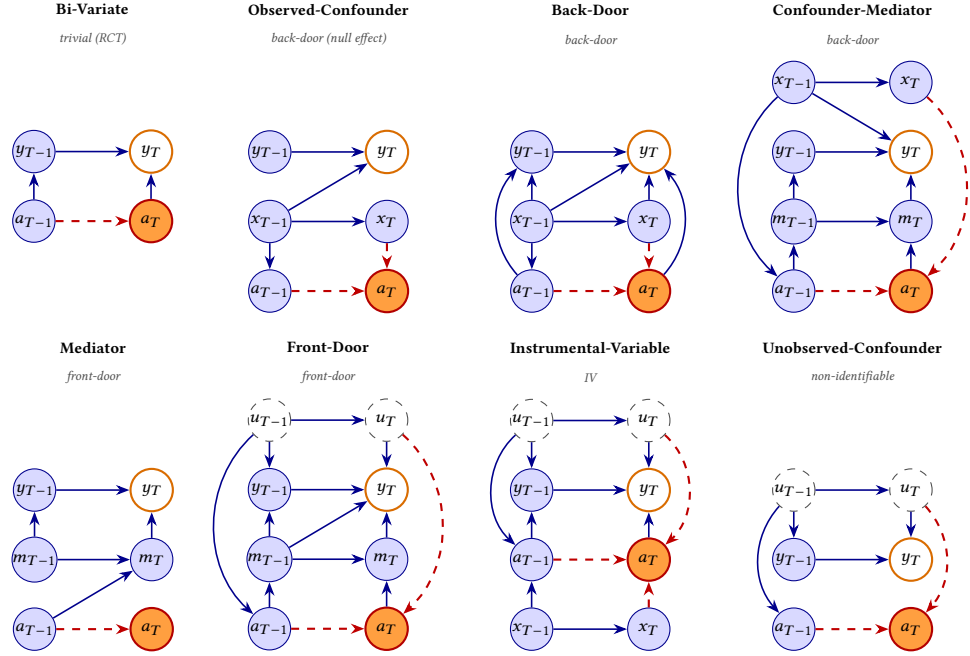
\begin{figure*}[t]
\centering
\begingroup
\tikzset{
  dtnode/.style = {circle, draw=blue!55!black, fill=blue!15, minimum size=5.6mm,
                   inner sep=0pt, font=\scriptsize},
  dtint/.style  = {circle, draw=red!70!black, thick, fill=orange!75, minimum size=5.6mm,
                   inner sep=0pt, font=\scriptsize},
  dtout/.style  = {circle, draw=orange!85!black, thick, fill=white, minimum size=5.6mm,
                   inner sep=0pt, font=\scriptsize},
  dthid/.style  = {circle, draw=black!70, dashed, fill=white, minimum size=5.6mm,
                   inner sep=0pt, font=\scriptsize},
  dtarr/.style  = {-{Stealth[length=1.8mm]}, blue!55!black, semithick},
  dtcut/.style  = {-{Stealth[length=1.8mm]}, red!75!black, thick, dashed},
  dttitle/.style= {font=\scriptsize\bfseries},
  dtfam/.style  = {font=\tiny\itshape, text=black!70},
}
\newcommand{\dtpanel}[3]{
  \begin{tikzpicture}[baseline=(current bounding box.north), x=1.55cm, y=0.92cm]
    \node[dttitle] at (0.5, 1.05) {#1};
    \node[dtfam]   at (0.5, 0.62) {#2};
    #3
  \end{tikzpicture}}
\setlength{\tabcolsep}{5pt}
\begin{tabular}{cccc}
\dtpanel{Bi-Variate}{trivial (RCT)}{%
  \node[dtnode] (a0) at (0,-2) {$a_{T-1}$}; \node[dtint] (a1) at (1,-2) {$a_{T}$};
  \node[dtnode] (y0) at (0,-1) {$y_{T-1}$}; \node[dtout] (y1) at (1,-1) {$y_{T}$};
  \draw[dtarr] (a0) -- (y0); \draw[dtarr] (a1) -- (y1);
  \draw[dtarr] (y0) -- (y1); \draw[dtcut] (a0) -- (a1);
}
&
\dtpanel{Observed-Confounder}{back-door (null effect)}{%
  \node[dtnode] (a0) at (0,-3) {$a_{T-1}$}; \node[dtint] (a1) at (1,-3) {$a_{T}$};
  \node[dtnode] (x0) at (0,-2) {$x_{T-1}$}; \node[dtnode] (x1) at (1,-2) {$x_{T}$};
  \node[dtnode] (y0) at (0,-1) {$y_{T-1}$}; \node[dtout] (y1) at (1,-1) {$y_{T}$};
  \draw[dtarr] (x0) -- (a0); \draw[dtcut] (x1) -- (a1);
  \draw[dtarr] (x0) -- (y1);
  \draw[dtarr] (x0) -- (x1); \draw[dtarr] (y0) -- (y1); \draw[dtcut] (a0) -- (a1);
}
&
\dtpanel{Back-Door}{back-door}{%
  \node[dtnode] (a0) at (0,-3) {$a_{T-1}$}; \node[dtint] (a1) at (1,-3) {$a_{T}$};
  \node[dtnode] (x0) at (0,-2) {$x_{T-1}$}; \node[dtnode] (x1) at (1,-2) {$x_{T}$};
  \node[dtnode] (y0) at (0,-1) {$y_{T-1}$}; \node[dtout] (y1) at (1,-1) {$y_{T}$};
  \draw[dtarr] (x0) -- (a0); \draw[dtcut] (x1) -- (a1);
  \draw[dtarr] (x0) -- (y0); \draw[dtarr] (x1) -- (y1);
  \draw[dtarr] (a0) to[bend left=42] (y0); \draw[dtarr] (a1) to[bend right=42] (y1);
  \draw[dtarr] (x0) -- (y1);
  \draw[dtarr] (x0) -- (x1); \draw[dtarr] (y0) -- (y1); \draw[dtcut] (a0) -- (a1);
}
&
\dtpanel{Confounder-Mediator}{back-door}{%
  \node[dtnode] (a0) at (0,-3) {$a_{T-1}$}; \node[dtint] (a1) at (1,-3) {$a_{T}$};
  \node[dtnode] (m0) at (0,-2) {$m_{T-1}$}; \node[dtnode] (m1) at (1,-2) {$m_{T}$};
  \node[dtnode] (y0) at (0,-1) {$y_{T-1}$}; \node[dtout] (y1) at (1,-1) {$y_{T}$};
  \node[dtnode] (x0) at (0,0)  {$x_{T-1}$}; \node[dtnode] (x1) at (1,0)  {$x_{T}$};
  \draw[dtarr] (x0) to[bend right=48] (a0); \draw[dtcut] (x1) to[bend left=48] (a1);
  \draw[dtarr] (a0) -- (m0); \draw[dtarr] (a1) -- (m1);
  \draw[dtarr] (m0) -- (y0); \draw[dtarr] (m1) -- (y1);
  \draw[dtarr] (x0) -- (y1);
  \draw[dtarr] (x0) -- (x1); \draw[dtarr] (m0) -- (m1);
  \draw[dtarr] (y0) -- (y1); \draw[dtcut] (a0) -- (a1);
}
\\[1.4em]
\dtpanel{Mediator}{front-door}{%
  \node[dtnode] (a0) at (0,-3) {$a_{T-1}$}; \node[dtint] (a1) at (1,-3) {$a_{T}$};
  \node[dtnode] (m0) at (0,-2) {$m_{T-1}$}; \node[dtnode] (m1) at (1,-2) {$m_{T}$};
  \node[dtnode] (y0) at (0,-1) {$y_{T-1}$}; \node[dtout] (y1) at (1,-1) {$y_{T}$};
  \draw[dtarr] (a0) -- (m1);
  \draw[dtarr] (m0) -- (y0); \draw[dtarr] (m1) -- (y1);
  \draw[dtarr] (m0) -- (m1); \draw[dtarr] (y0) -- (y1); \draw[dtcut] (a0) -- (a1);
}
&
\dtpanel{Front-Door}{front-door}{%
  \node[dtnode] (a0) at (0,-3) {$a_{T-1}$}; \node[dtint] (a1) at (1,-3) {$a_{T}$};
  \node[dtnode] (m0) at (0,-2) {$m_{T-1}$}; \node[dtnode] (m1) at (1,-2) {$m_{T}$};
  \node[dtnode] (y0) at (0,-1) {$y_{T-1}$}; \node[dtout] (y1) at (1,-1) {$y_{T}$};
  \node[dthid] (u0) at (0,0)  {$u_{T-1}$}; \node[dthid] (u1) at (1,0)  {$u_{T}$};
  \draw[dtarr] (u0) to[bend right=48] (a0); \draw[dtcut] (u1) to[bend left=48] (a1);
  \draw[dtarr] (u0) -- (y0); \draw[dtarr] (u1) -- (y1);
  \draw[dtarr] (a0) -- (m0); \draw[dtarr] (a1) -- (m1);
  \draw[dtarr] (m0) -- (y0); \draw[dtarr] (m1) -- (y1);
  \draw[dtarr] (m0) -- (y1);
  \draw[dtarr] (u0) -- (u1); \draw[dtarr] (m0) -- (m1);
  \draw[dtarr] (y0) -- (y1); \draw[dtcut] (a0) -- (a1);
}
&
\dtpanel{Instrumental-Variable}{IV}{%
  \node[dtnode] (x0) at (0,-3) {$x_{T-1}$}; \node[dtnode] (x1) at (1,-3) {$x_{T}$};
  \node[dtnode] (a0) at (0,-2) {$a_{T-1}$}; \node[dtint] (a1) at (1,-2) {$a_{T}$};
  \node[dtnode] (y0) at (0,-1) {$y_{T-1}$}; \node[dtout] (y1) at (1,-1) {$y_{T}$};
  \node[dthid] (u0) at (0,0)  {$u_{T-1}$}; \node[dthid] (u1) at (1,0)  {$u_{T}$};
  \draw[dtarr] (x0) -- (a0); \draw[dtcut] (x1) -- (a1);
  \draw[dtarr] (a0) -- (y0); \draw[dtarr] (a1) -- (y1);
  \draw[dtarr] (u0) -- (y0); \draw[dtarr] (u1) -- (y1);
  \draw[dtarr] (u0) to[bend right=48] (a0); \draw[dtcut] (u1) to[bend left=48] (a1);
  \draw[dtarr] (x0) -- (x1); \draw[dtarr] (u0) -- (u1);
  \draw[dtarr] (y0) -- (y1); \draw[dtcut] (a0) -- (a1);
}
&
\dtpanel{Unobserved-Confounder}{non-identifiable}{%
  \node[dtnode] (a0) at (0,-3) {$a_{T-1}$}; \node[dtint] (a1) at (1,-3) {$a_{T}$};
  \node[dtnode] (y0) at (0,-2) {$y_{T-1}$}; \node[dtout] (y1) at (1,-2) {$y_{T}$};
  \node[dthid] (u0) at (0,-1) {$u_{T-1}$}; \node[dthid] (u1) at (1,-1) {$u_{T}$};
  \draw[dtarr] (u0) -- (y0); \draw[dtarr] (u1) -- (y1);
  \draw[dtarr] (u0) to[bend right=42] (a0); \draw[dtcut] (u1) to[bend left=42] (a1);
  \draw[dtarr] (u0) -- (u1); \draw[dtarr] (y0) -- (y1); \draw[dtcut] (a0) -- (a1);
}
\end{tabular}
\endgroup

\caption{All eight identification structures of \suite{dot-Identifiability-v1} as \emph{temporal} SCMs, unrolled over two consecutive slices $T{-}1$ (pre-intervention) and $T$ (intervened). The identification-strategy taxonomy follows~\cite{thumm2026causal}, with edges as instantiated by the released generator (\code{tscm\_sampler}, lagged variant). Solid blue arrows are causal edges, either  instantaneous within a slice, or lagged across slices, including an autoregressive lag-$1$ self-edge on every variable. Each structure additionally carries a lag that makes its identification strategy visible from history (e.g.\ $x_{T-1}{\to}y_T$ for the back-door family, $m_{T-1}{\to}y_T$ for front-door). The intervention $\mathrm{do}(a_T)$ (orange node) severs the incoming edges of $a_T$ (red dashed) while its outgoing edges stay active; the queried outcome $y_T$ is the open node, and dashed nodes ($u$) are hidden. Identification status per structure is in Table~\ref{tab:structures}.}
\label{fig:structures}
\end{figure*}

\section{Scope and Extensions}\label{sec:scope}

The four axes below are DoTime's main extensions beyond prior generators. Each ships a concrete API and is exercised by a released suite.

\subsection{Intervention types}
\begin{figure}[t]
\centering
\includegraphics[width=\columnwidth]{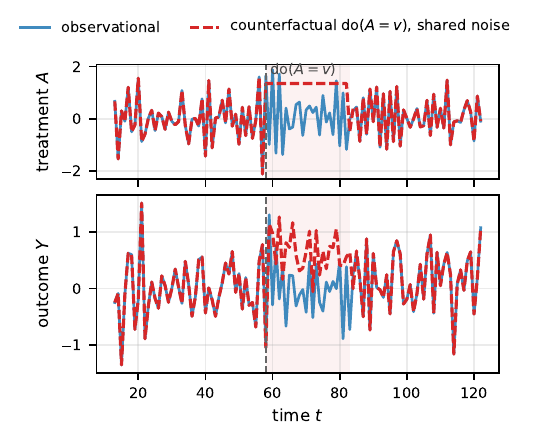}
\caption{One paired trajectory from the continuous-time back-door generator (\suite{dot-Continuous-v1}); both arms share one noise realisation. 
Top: the treatment $A$, clamped to $v$ inside the intervention window (shaded). 
Bottom: the outcome $Y$. Pre-onset the counterfactual arm (red, dashed) is \emph{bit-identical} to the observational arm (blue). It diverges only inside the window and re-synchronizes after it ends as the mean-reverting dynamics forget the clamp. The visible difference \emph{is} the causal effect --- a counterfactual, not an independent interventional draw.} 
\label{fig:traj}
\end{figure}

We support \emph{hard} $\mathrm{do}(A_t=v)$, \emph{soft} (an additive shift to the mechanism output), and \emph{time-varying} $\mathrm{do}(A_t = v(t))$ interventions, encoded as a single \code{InterventionSpec} so the same SCM can be
queried under all three. Figure~\ref{fig:traj} shows a paired observational / interventional trajectory from the \emph{continuous-time} generator, which simulates both arms under one shared noise realisation: pre-onset the two runs are bit-identical, inside the window the clamped treatment drives $Y$ off its observational path, and after the window the arms re-synchronize as the dynamics forget the clamp. Every visible difference is caused by the intervention alone. The discrete-time generators instead pair each observational trajectory with an \emph{independent} interventional draw from the same SCM which relates to interventional rather than
counterfactual ground truth (\S\ref{sec:limitations}).

\subsection{Continuous-time intervention windows}
\begin{figure}[t]
\centering
\includegraphics[width=0.66\columnwidth]{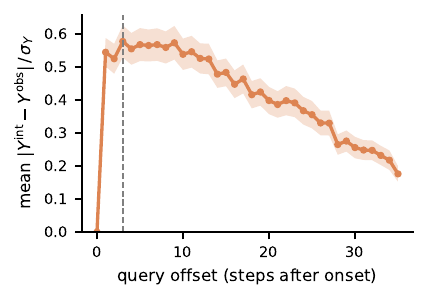}
\caption{Mean absolute intervention effect on the outcome, in units of the outcome's own standard deviation ($\sigma_Y$), vs.\ query offset, averaged over back-door TSCMs from the shared-noise continuous generator (band: $\pm1$ SE). Because the two arms share one noise realisation, the difference is the causal effect alone --- it builds during the intervention window, then decays as the mean-reverting dynamics forget the clamp, justifying the multi-offset query protocol.}
\label{fig:decay}
\end{figure}

We generalize a single-step intervention to a \emph{window} $\mathrm{do}(A_{[t_0,t_1]}=v)$, encoded by floating-point \code{intervention\_time\_start} and \code{intervention\_time\_end} in $[0,1]$.
Continuous-time trajectories are generated by Euler--Maruyama integration of an underlying stochastic differential equation (SDE) on a fine grid, of which the observed series is a discretization~\cite{kloeden1992numerical}. This lets us vary window duration independently of the observation cadence.
Crucially, the observation schedule itself is configurable --- \emph{regular}, \emph{jittered} ($\Delta_i = \bar\Delta(1+\xi_i)$, $\xi_i\!\sim\!\mathrm{Unif}[-\rho,\rho]$), or a \emph{Poisson} point process ($\Delta_i\!\sim\!\mathrm{Exp}(1/\bar\Delta)$) --- so the generator natively covers genuinely irregular, non-uniformly spaced observation times rather than deferring them to future work. Figure~\ref{fig:decay} shows that the mean intervention effect on the outcome decays with the query offset, motivating the multi-offset evaluation protocol.

\subsection{Counterfactual sampling modes}
The intervention value can be drawn from five modes: an unguarded
extrapolative draw, the same draw under a positivity guard, and three
in-support modes. The guard clips to $[\hat\mu\pm 3\hat\sigma]$, estimated from
the pre-intervention window of the treated variable
(\code{positivity\_aware} $=$ \code{prior} $+$ clip):
\begin{align*}
\text{\code{prior}}: & \;\; v \sim p_{\text{prior}} \quad\text{(extrapolative)},\\
\text{\code{observed\_discrete}}: & \;\; v \sim \mathrm{Unif}\{A_{<t_0}\},\\
\text{\code{observed\_normal}}: & \;\; v \sim \mathcal{N}(\hat\mu,\hat\sigma^2),\\
\text{\code{observed\_uniform}}: & \;\; v \sim \mathrm{Unif}[\min A_{<t_0}, \max A_{<t_0}].
\end{align*}
This choice governs a distinction users repeatedly face: whether a model is
trained on extrapolative interventions or on counterfactuals inside the
observed support. We
quantify the resulting out-of-distribution (OOD) ratio and effect magnitude per mode in
Appendix~\ref{app:hparams}.

\subsection{Regime-switching as an ITS generalization}
We define a sticky-Markov regime chain over TSCM parameters with a high self-transition probability. A two-regime chain with a deterministic transition at $t^\star$ recovers the piecewise interrupted-time-series (ITS) design --- a single \emph{known} changepoint shifting both level and dynamics, exactly the step-plus-slope change a segmented regression specifies, where a correctly specified ITS estimator is unbiased and the exact ground truth lets the benchmark verify this rather than assume it. Relaxing those assumptions one at a time --- stochastic switch times, more than two regimes, nonlinear within-regime mechanisms --- breaks the segmented-regression specification while leaving the ground truth exact, so the same suite stress-tests classical Bayesian piecewise ITS~\cite{causalpy2026} \emph{and} modern causal foundation models head-to-head. The released suite varies regime density over $\{2,3,5\}$, exposed as a difficulty tier.

\section{The Released Benchmark Suites}\label{sec:suites}

We release four versioned, immutable suites. Each is stored as parquet shards plus a checksummed \code{manifest.json} and Croissant JSON-LD, and every episode carries a difficulty \emph{tier}.

\paragraph{\suite{dot-Identifiability-v1}.} Eight named identification structures --- \code{back\_door}, \code{observed\_confounder}, \code{confounder\_mediator} (back-door family); \code{front\_door}, \code{mediator} (front-door family); \code{instrumental\_variable}; \code{bi\_variate} (trivially identified); and \code{unobserved\_confounder} (non-identifiable by design, testing graceful degradation). Among these, \code{observed\_confounder} is a deliberate \emph{null-effect} control: $X$ confounds $A$ and $Y$ but $A$ has no causal path to $Y$, so the confounded observational correlation is spurious and back-door adjustment must recover a zero effect. $\approx\!1{,}350$ episodes per structure ($\approx\!10.8$k total), $T\!=\!200$, with exact interventional targets and a per-structure intervention/query offset protocol matched to identification theory. 
We state the identification asymmetry plainly~\cite{pearl2009causality}: back-door and front-door are nonparametrically point-identifiable. The instrumental-variable case identifies only under linearity/monotonicity. The unobserved-confounder case is non-identifiable by design, included to measure graceful degradation rather than accuracy. Table~\ref{tab:structures} lists the eight structures and their identification status\footnote{``Bi-Variate'' (\code{bi\_variate}) is the trivial $A\to Y$ design --- an in-silico randomized controlled trial (RCT) with no confounding. The archived v1.0.0 files carry this structure's legacy label \code{rct\_no\_confounding}, which \code{load\_benchmark} relabels to \code{bi\_variate} on read.}.

\begin{table}[t]
\caption{The eight canonical identification structures in \suite{dot-Identifiability-v1}, by family and identification status. Display names map one-to-one to the code identifiers in \S\ref{sec:suites}.}
\label{tab:structures}
\small
\setlength{\tabcolsep}{4pt}
\begin{tabular}{lll}
\toprule
Structure & Family & Identifiable? \\
\midrule
Bi-Variate            & trivial (RCT)   & \cmark{} (direct) \\
Observed-Confounder   & back-door       & \cmark{} (null effect) \\
Back-Door             & back-door       & \cmark \\
Confounder-Mediator   & back-door       & \cmark \\
Mediator              & front-door      & \cmark \\
Front-Door            & front-door      & \cmark \\
Instrumental-Variable & IV              & \cmark{} (linear/mono.) \\
Unobserved-Confounder & ---             & \xmark{} (by design) \\
\bottomrule
\end{tabular}
\end{table}

\paragraph{\suite{dot-RegimeSwitch-v1}.} $10$k trajectories, regime density $\in\{2,3,5\}$; designed as the target for CausalPy-style~\cite{causalpy2026} piecewise-ITS baselines.

\paragraph{\suite{dot-Continuous-v1}.} $10$k trajectories with continuous-time intervention windows of varying duration and query offsets $\{1,2,3,5,10\}$.

\paragraph{\suite{dot-Generic-100k}.} $100{,}000$ trajectories from the full diverse prior. It is an order of magnitude larger than the other three because it serves a different purpose: the diagnostic suites isolate a named structure or mechanism and reach tight error bars at $\approx\!10$k episodes, whereas this one is the \emph{training-scale} snapshot, sized for pre-training a foundation model rather than for scoring one. It is validated for effect-magnitude distribution and intervention-type ratio. See \S\ref{sec:validation} for the divergence-handling disclosure.

\paragraph{Distribution and reproducibility.} Each suite is archived on Zenodo with a citable DOI and mirrored as a Hugging Face dataset for discovery. \code{load\_benchmark} pulls from either and verifies md5 checksums against the manifest. We cite each suite's version-independent \emph{concept} DOI, which always resolves to its latest archived version: \texttt{10.5281/zenodo.20846063} (Identifiability), \texttt{.20846073} (RegimeSwitch), \texttt{.20845980} (Continuous), and \texttt{.20845982} (Generic-100k); the trained Do-Over-Time-PFN checkpoints are released on the Hugging Face Hub. Every suite regenerates from the committed \code{scripts/build\_release.py} with fixed seeds, recording package version, config hash, and hardware in a build manifest. We adhere to the ACM \emph{Artifacts Available} badge standard: all data, code, and model weights are public under Apache-2.0 (code) / CC-BY-4.0 (data) with persistent DOIs.

\section{Baselines and Reference Results}\label{sec:baselines}

\subsection{Protocol}
We predict the interventional outcome $Y^{\text{int}}_{t+\text{offset}}$ on its raw (un-normalized) scale, with per-trajectory normalization applied at evaluation time and bootstrap confidence intervals resampled over trajectories. We bootstrap rather than quote a closed-form interval because pooled RMSE aggregates several correlated queries within each episode: treating those queries as independent would understate the true uncertainty, so we resample whole episodes (the independent unit) instead. Near-zero targets ($|Y|<0.1$) are excluded from direction accuracy and reported separately.

\subsection{Baselines}
The package registers six baselines evaluated end-to-end in Table~\ref{tab:results}: \code{Zero} (a constant zero prediction, a trivial sanity floor); trajectory heuristics (\code{TrajMean}, the pre-intervention outcome mean; \code{AR1}, the last observed value); a vector autoregression (\code{VAR}); and classical structural adjustment (\code{BackDoorOLS}, a linear back-door adjustment; \code{IV2SLS}, two-stage least squares with a weak-instrument guard). A covariate-informed observational forecaster (\code{Chronos-2}~\cite{ansari2025chronos2}) requires optional dependencies and is evaluated out-of-band in \S\ref{sec:fmgap}; a CausalPy~\cite{causalpy2026} piecewise-ITS baseline is a natural fit for \suite{dot-RegimeSwitch-v1} and is left as an explicitly supported entry point for external submissions rather than a baseline we report here. Alongside these we report the headline \textbf{Do-Over-Time-PFN}~\cite{thumm2026causal} causal foundation model in two training conditions of identical capacity, interventional (\code{PFN\textsubscript{int}}) and observational (\code{PFN\textsubscript{obs}}). \code{Oracle} reads off the released ground-truth target, it is an upper bound by construction on the synthetic suites. Direction accuracy scores the sign of the predicted effect against ground truth; a constant prediction (\code{Zero}) has no sign and is assigned $0$ by convention.

\subsection{Results}\label{sec:results}
Table~\ref{tab:results} reports pooled RMSE and direction accuracy across all four suites. On these nonlinear TSCMs every learned baseline sits far above the \code{Oracle} (RMSE floor $0$, accuracy ceiling $1$): RMSE is hard to beat because the per-trajectory series cluster around their means. Squared-error metrics are in fact near-saturated. In normalized MSE (NMSE $=$ MSE$/\mathrm{Var}(Y)$, so $1.0$ is the predict-the-mean floor), the best Identifiability baseline reaches only $0.75$ and every learned method sits in $[0.75, 1.52]$ --- absolute error cannot rank causal quality here. This is a property of the task, not a call for a sharper regression metric: when every model is near the predict-the-mean floor, squared error is dominated by the irreducible noise rather than by how well a model answers the $\mathrm{do}$-query. It is why our analysis leads with \emph{direction accuracy} and the interventional-vs-observational \emph{gap} (\S\ref{sec:fmgap}, \S\ref{sec:gap}): those isolate causal understanding at fixed capacity, where RMSE cannot. Closing the remaining gap to the \code{Oracle} is then a modelling problem, but one the benchmark can only \emph{score} through the causal metrics, not through RMSE.

We state the uncomfortable consequence plainly: on direction accuracy, \code{PFN\textsubscript{int}} does \emph{not} beat \code{TrajMean} on Identifiability ($0.66$ vs.\ $0.67$) and ties it on Generic ($0.66$). A model that commits to an intervention direction pays a small variance penalty relative to a mean predictor, and no absolute comparison against \code{TrajMean} in Table~\ref{tab:results} favors the PFN. The benchmark's discriminating signal is instead the \emph{contrast}: the interventional model against its observational twin at \emph{identical capacity}, where every confound except interventional training cancels.

That contrast behaves as causal reasoning predicts. On the two suites closest to the checkpoint's training distribution the interventional PFN lifts direction accuracy over its observational twin ($0.66$ vs.\ $0.57$ on Identifiability,
$+0.08$; $0.66$ vs.\ $0.62$ on Generic, $+0.04$); on Regime and Continuous whose dynamics this checkpoint never saw in any form. The advantage washes out (gaps $-0.01$ and $+0.02$), an honest boundary on what a single discrete-prior model transfers to. \S\ref{sec:fmgap} shows this gap is a property of interventional \emph{training}, not test-time access to the intervention value, and the structure-matched analysis in \S\ref{sec:gap} isolates it cleanly.

\begin{table*}[t]
\caption{Reference results at \textbf{release scale} (full suites:
$10{,}800$ / $9{,}999$ / $9{,}999$ / $100{,}000$ episodes): pooled RMSE (direction
accuracy in parentheses); \code{Oracle} is the synthetic upper bound. PFN rows use
one interventional and one observational checkpoint of identical capacity
(\code{s9ho\_all\_causal} / \code{s9ho\_all\_obs}; \code{OSC} prior, no lagged
edges --- no suite matches this training distribution exactly, see \S\ref{sec:results}).
\code{BackDoorOLS}/\code{IV2SLS} fall back to the outcome mean off their
applicable structures; Generic RMSE is on the raw scale. RMSE cells show
mean\,$\pm$\,std (bootstrap $95\%$ CI width\,/\,$3.92$); direction-accuracy
standard errors ($\sqrt{p(1-p)/n}$) are $\le\!0.01$ on every full-suite row.
\gapnote}
\label{tab:results}
\small
\setlength{\tabcolsep}{10pt}
\begin{tabular}{lcccc}
\toprule
Baseline & Ident. & Regime & Cont. & Generic \\
\midrule
Oracle (upper bnd)      & 0.00 (1.00) & 0.00 (1.00) & 0.00 (1.00) & 0.00 (1.00) \\
Zero                    & $0.72\sd{.02}$ (0.00) & $1.82\sd{.04}$ (0.00) & $2.75\sd{.61}$ (0.00) & $46.2\sd{1.4}$ (0.00) \\
TrajMean                & $0.62\sd{.01}$ (0.67) & $1.84\sd{.03}$ (0.50) & $2.74\sd{.61}$ (0.50) & $46.6\sd{1.4}$ (0.66) \\
AR1                     & $0.74\sd{.02}$ (0.60) & $2.54\sd{.05}$ (0.49) & $3.40\sd{.67}$ (0.51) & $49.4\sd{1.5}$ (0.62) \\
VAR                     & $0.66\sd{.01}$ (0.64) & $1.92\sd{.04}$ (0.50) & $3.50\sd{.56}$ (0.50) & $62.1\sd{4.6}$ (0.65) \\
BackDoorOLS             & $0.78\sd{.02}$ (0.65) & $1.84\sd{.03}$ (0.50) & $2.77\sd{.60}$ (0.53) & $46.6\sd{1.4}$ (0.66) \\
IV2SLS                  & $0.88\sd{.06}$ (0.67) & $1.84\sd{.03}$ (0.50) & $2.76\sd{.60}$ (0.51) & $46.6\sd{1.4}$ (0.66) \\
PFN\textsubscript{obs}  & $0.67\sd{.01}$ (0.57) & $2.07\sd{.04}$ (0.51) & $2.97\sd{.70}$ (0.49) & $46.3\sd{1.4}$ (0.62) \\
PFN\textsubscript{int}  & $0.67\sd{.01}$ (0.66) & $1.99\sd{.04}$ (0.49) & $4.07\sd{.87}$ (0.51) & $48.2\sd{1.4}$ (0.66) \\
\bottomrule
\end{tabular}
\end{table*}

\subsection{Training vs.\ test-time context}\label{sec:fmgap}
A skeptic could object that the interventional PFN wins simply because it \emph{sees} the intervention value at test time. We rule this out by giving the same do-value to two observational foundation models and measuring their interventional-vs-observational direction-accuracy gap on \suite{dot-Identifiability-v1} (Table~\ref{tab:fmgap}). \textbf{TabPFN}~\cite{tabpfn2023} runs a back-door / front-door adjustment with the do-value plugged into two (resp.\ three) in-context regressors. \textbf{Chronos-2}~\cite{ansari2025chronos2} forecasts the outcome with the intervened treatment supplied as a known future covariate. Neither observational model shows a gap ($-0.01$, $-0.01$): access to the do-value at test time buys nothing. Only the DoTime-\emph{trained} PFN separates the two conditions ($+0.08$, well outside its standard error, while both observational models' gaps sit within theirs). TabPFN in fact posts the highest absolute interventional accuracy of the three while showing no gap at all --- absolute levels and the gap measure different things, and the claim under test concerns the gap. The advantage is a property of interventional pre-training, not of conditioning on the intervention at inference\footnote{\gapnote}.

\begin{table}[t]
\caption{Interventional vs.\ observational direction accuracy
($\pm$ standard error) on \suite{dot-Identifiability-v1}, all three models on the \emph{same} $480$-episode stratified subsample ($60$ per structure; TabPFN and Chronos-2 are too costly per episode for the full suite). Bold marks the gap
column --- the within-row contrast; absolute accuracies are not comparable across rows (the models differ in capacity, pre-training corpus, and adjustment machinery). The PFN's gap is unchanged on the full suite ($0.66$ vs.\ $0.57$, Table~\ref{tab:results}), so the subsample is not driving it.}
\label{tab:fmgap}
\small
\setlength{\tabcolsep}{5pt}
\begin{tabular}{lccc}
\toprule
Model & interventional & observational & gap \\
\midrule
TabPFN (adjustment)     & $0.67\sd{.02}$ & $0.68\sd{.02}$ & $-0.01$ \\
Chronos-2               & $0.62\sd{.03}$ & $0.63\sd{.02}$ & $-0.01$ \\
\textbf{Do-Over-Time-PFN} & $0.64\sd{.02}$ & $0.55\sd{.03}$ & $\mathbf{+0.08}$ \\
\bottomrule
\end{tabular}
\end{table}

\subsection{The observational-vs-interventional gap}\label{sec:gap}
Table~\ref{tab:gap} is the central contrast and the test of our falsifiable claim. We use two hardened priors: code{OSC}, an \emph{oscillatory} prior, and \code{BTM}, a \emph{break-trajectory-mean} prior whose trajectories deliberately depart from their running mean (so a mean-predictor cannot succeed by default).
For each of the three point-identified structures (back-door, front-door, instrumental-variable; an exploratory probe of two further structures is in Appendix~\ref{app:perstruct}) we evaluate two structure-matched Do-Over-Time-PFN models of \emph{identical capacity} --- one trained on interventional context, one on observational context only. 
Both models are trained and evaluated on the \emph{same} prior configuration. What is held out is the \emph{episodes} (training and evaluation use disjoint random seeds), so this measures the interventional-training advantage in-distribution, not under generator shift. 
A further protocol fact matters for interpretation: the primary study trains and evaluates \emph{without lagged edges} (instantaneous-graph dynamics only, \code{--no-tscm-lag}). A lagged replication (identical protocol with lagged edges enabled in both training and evaluation, single seed) preserves the gap on every structure: back-door $+0.12\sd{.03}$, front-door $+0.04\sd{.02}$, instrumental-variable $+0.06\sd{.02}$, pooled $+0.07$ (lagged) vs.\ $+0.08$ (lag-free, same $n{=}960$ evaluation batch; the Table~\ref{tab:gap} sweep pools to $+0.09$ at $T{=}200$ on its smaller $n{=}640$ batch). The advantage is thus not an artifact of the instantaneous-graph setting. Per-cell standard errors capture evaluation sampling noise; training-seed variance is quantified separately below.

To rule out trajectory length as a confounder we sweep $T \in \{200, 500, 1000, 2000\}$ (Table~\ref{tab:gap}). The interventional model attains a higher direction accuracy for every structure at every length. On the \code{OSC} prior the gap reaches $+0.17$ on the back-door case and pools to $\approx\!+0.09$, essentially flat across a $10\times$ range of trajectory lengths. On the \code{BTM} prior the trend is the same but weaker (pooled $+0.02$ to $+0.05$ across $T$, Appendix~\ref{app:perstruct}). RMSE is similar across the two models, so the signal is in the \emph{sign} of the estimated effect: on the back-door case the observational model recovers the direction $0.63$ of the time vs.\ $0.78$ for the interventional model ($T{=}200$, Table~\ref{tab:gap} source run).

To rule out training-seed luck --- the variance the per-cell standard errors do \emph{not} cover --- we retrain every \code{OSC} arm from scratch under three seeds and re-evaluate at $T{=}200$. The gap is positive in all nine structure$\times$seed cells; per-structure means $\pm$ seed std are back-door $+0.14\sd{.02}$, front-door $+0.04\sd{.01}$, instrumental-variable $+0.09\sd{.04}$, and the pooled gap is $+0.089$ with a seed-to-seed std of only $0.005$ (per-seed pooled values $+0.084$, $+0.090$, $+0.094$). Across three training seeds, four trajectory lengths, and three structures, interventional training buys a real and measurable advantage that the benchmark detects\footnote{\tgapnote\ \gapnote}. Figure~\ref{fig:cid} (Appendix~\ref{app:perstruct}) shows the mechanism on one representative SCM per structure: the interventional model's predictive density shifts toward the ground-truth conditional interventional distribution while its observational twin stays on the observational density --- and the visual separation is ordered exactly as the measured gaps, largest for back-door and smallest for IV.

\begin{table}[t]
\caption{Direction-accuracy gap (int $-$ obs, $\pm$ standard error) of the structure-matched interventional vs.\ observational Do-Over-Time-PFN
on the \code{OSC} prior, evaluated on held-out episodes (disjoint seeds; same prior configuration as training; no lagged edges), across trajectory lengths $T \in \{200, 500, 1000, 2000\}$.}
\label{tab:gap}
\small
\setlength{\tabcolsep}{3pt}
\begin{tabular}{lcccc}
\toprule
Structure & $T{=}200$ & $T{=}500$ & $T{=}1000$ & $T{=}2000$ \\
\midrule
back\_door             & $+0.14\sd{.03}$ & $+0.12\sd{.03}$ & $+0.17\sd{.03}$ & $+0.17\sd{.03}$ \\
front\_door            & $+0.08\sd{.03}$ & $+0.07\sd{.03}$ & $+0.07\sd{.03}$ & $+0.07\sd{.03}$ \\
instrumental\_variable & $+0.05\sd{.03}$ & $+0.04\sd{.03}$ & $+0.03\sd{.03}$ & $+0.04\sd{.03}$ \\
\midrule
\textbf{pooled}        & $\mathbf{+0.09}$ & $\mathbf{+0.08}$ & $\mathbf{+0.09}$ & $\mathbf{+0.10}$ \\
\bottomrule
\end{tabular}
\end{table}

\section{Representativeness and Validation}\label{sec:validation}

\subsection{Representativeness: zero-shot transfer to real systems}\label{sec:repr}
\begin{figure*}[t]
\centering
\includegraphics[width=0.92\textwidth]{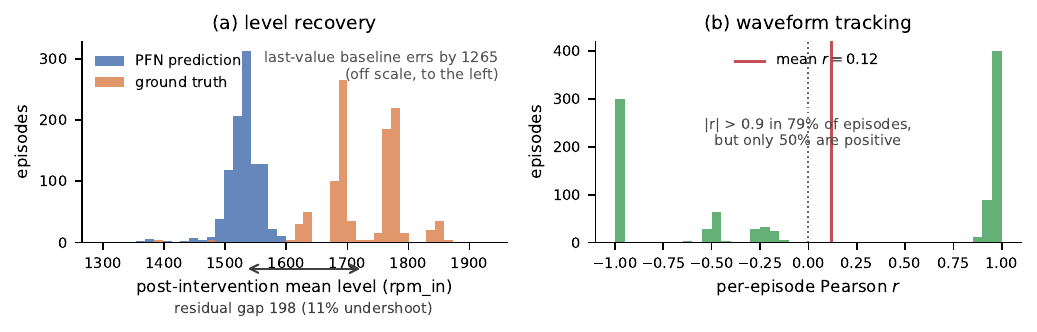}
\caption{Level recovery versus waveform tracking on the Causal Chambers \code{wt\_intake\_impulse} probe ($1{,}000$ episode$\times$seed evaluations, five training seeds). \textbf{(a)}~After the impulse the synthetic-trained PFN places the operating level within $198$ of ground truth, against the $1265$ by which a last-value baseline (anchored at the pre-intervention reading, off scale to the left) misses --- but it undershoots systematically, and its predicted level is nearly constant while the true level varies. \textbf{(b)}~Within the post-intervention window the per-episode correlation is bimodal at $\pm1$: the model tracks a strongly-shaped curve, but its phase is close to a coin flip.
Together the panels are why a large RMSE lift and a near-zero mean correlation are consistent rather than contradictory.} 
\label{fig:levelwave}
\end{figure*}

For a synthetic data generator, representativeness is the load-bearing requirement:
do models trained \emph{only} on the prior behave sensibly on real dynamics? We answer by testing external validity: zero-shot transfer to real systems. A causal foundation model trained purely on DoTime-style synthetic TSCMs transfers \textbf{zero-shot} (no fine-tuning) to two disjoint real domains~\cite{thumm2026towards}.
\paragraph{Physical --- Causal Chambers wind tunnel.} On the \code{wt\_intake\_impulse} benchmark~\cite{causalchamber2024} (predicting \code{rpm\_in} under an intake impulse), the synthetic-trained PFN attains RMSE $661\!\pm\!4$ vs.\ a naive last-value baseline at $1265$ --- a \textbf{$+47.8\%\!\pm\!0.3\%$ lift} over five training seeds --- yet its per-episode Pearson correlation is near zero ($0.12\!\pm\!0.17$). The two are consistent because they measure different things: the model relocates the post-intervention \emph{operating level} (with a systematic $11\%$ undershoot in $99\%$ of episodes) but does not track the chamber's high-frequency micro-oscillations, whose sign flips faster than it can follow. Figure~\ref{fig:levelwave} makes the split visible and quantifies both failures; the representativeness signal is regime relocation, not waveform tracking. 
At panel~(a) the predicted and true level distributions sit close together on the scale that matters. The residual gap between them is $198$, against the $1265$ a last-value baseline is off by. 
Meanwhile, panel~(b) shows the correlation is bimodal at $\pm 1$, with $|r|>0.9$ in $79\%$ of episodes but only $50\%$ of them positive. The model therefore produces a strongly-shaped curve whose \emph{phase} is close to a coin flip. One further limit is worth stating: within this probe the predicted level is nearly constant ($1530\!\pm\!28$) while the true level varies more ($1728\!\pm\!63$), so across episodes the two are uncorrelated ($r\!=\!-0.01$); the model relocates the operating point to the right \emph{regime} rather than tracking each episode's level. In short, it recovers the intervention's \emph{coarse magnitude and level} but neither the per-episode level nor the fine \emph{waveform} --- the representativeness signal here is regime relocation, not oscillation tracking.
\paragraph{Biological --- pharmacokinetics.} On real Warfarin \\ plasma-concentration trajectories\footnote{First-order compartmental dosing --- the standard clinical PK model --- with dose as the intervention; a mixed-mechanism variant was also run and is weaker/more seed-unstable (see \code{results/reference/transfer/}).}, the same model family co-varies with the dose-driven trajectory: four of five seeds attain \textbf{Pearson $r\!\in\![0.72, 0.89]$} zero-shot, though the fifth collapses to $r\!\approx\!-0.87$. The correlation is real but seed-unstable, and the two halves of that statement need different evidence. Real: pooling the $160$ per-subject correlations gives mean $0.49$, median $0.87$, with $80\%$ positive --- significant against zero by a one-sample $t$-test ($t\!=\!8.7$, $p\!<\!10^{-4}$), a Wilcoxon signed-rank test, and a sign test ($p\!<\!10^{-4}$ each). Seed-unstable: the same claim tested at the \emph{seed} level, on five per-seed means, does \emph{not} reach significance ($t\!=\!1.44$, $p\!=\!0.22$) --- one collapsed seed is enough to erase it. We report both tests rather than the flattering one, since the gap between them is precisely the instability we are disclosing. On error magnitude the probe is close to null: RMSE $3.49\!\pm\!0.04$ against a naive last-value baseline at $3.51$, a lift of $0.5\%\!\pm\!1.1\%$ --- far below the chamber's $+47.8\%$.

We report these as five-seed transfer probes, not competitive benchmarks: the RMSE lift on the pharmacokinetic (PK) probe is negligible for the reason just given (the per-subject series cluster tightly around their means, so a last-value baseline is already near-optimal and there is little headroom to win) and the chamber signal is level recovery rather than dynamics tracking. We claim representativeness --- synthetic-trained models behave sensibly on real interventional dynamics --- not competitive dominance over domain-specific tools. Purely synthetic pretraining is, moreover, no longer an outlier position in time series: TabPFN-TS-3 is trained on synthetic series alone yet ranks second among time-series foundation models on fev-bench~\cite{grinsztajn2026tabpfn3}, and Chronos-2 derives its multivariate and covariate-informed capabilities entirely from synthetic data, with a synthetic-only variant nearly matching its full model~\cite{ansari2025chronos2}.
The per-seed transfer metrics behind these numbers ship under \code{results/reference/transfer/}; the transfer scatter and full tables are in \citet{thumm2026towards}.

\subsection{Internal validation and sanity checks}
\paragraph{Acyclicity and stability.} A diverged trajectory is released as an all-zero episode with a zero target, excluded from direction-accuracy scoring by the near-zero rule of the evaluation protocol. In the released v1.0.0 suites the zeroed fraction is $28.7\%$ on \suite{dot-Generic-100k} and $4.6\%$ on \suite{dot-Identifiability-v1} ($0\%$ on RegimeSwitch / Continuous) --- higher than we consider acceptable at the generic prior's default configuration, and disclosed here so users can filter on $\lVert X\rVert > 0$. Appendix~\ref{app:theory} traces the cause to the deliberately relaxed spectral-radius assumption and describes the opt-in fix (deterministic resampling; $0/200$ zeroed on a full generic build, regression-tested).
\paragraph{Identifiability sanity.} \code{Oracle} attains RMSE $\approx 0$ on every synthetic suite (asserted in our test suite).
\paragraph{Distribution shift between suites.} Summary statistics of \suite{dot-RegimeSwitch-v1} and \suite{dot-Generic-100k} differ non-trivially, justifying separate evaluation.
\paragraph{Counterfactual modes.} The \code{observed\_normal} mode yields strictly smaller observational-baseline errors than \code{prior}, confirming the positivity advantage.

\section{Responsible Release and Limitations}\label{sec:limitations}

\paragraph{Limitations.} The \emph{frozen suites} use $N\!\le\!10$ variables and $K\!\le\!3$ lags; these are the \code{DEFAULT\_CONFIG} bounds (\code{N\_max}, \code{K\_max}) we chose for the released snapshots, not architectural limits. The same generator produces larger graphs and deeper lags from a config override (a regression test in the released package generates at $N_{\max}\!=\!60$, $K_{\max}\!=\!8$), so scale is a user choice rather than a ceiling. Divergence rates grow with scale absent the hardening configuration (\S\ref{sec:validation}). Non-stationarity enters the suites through switching SCM parameters and time-varying intervention profiles, but the \emph{exogenous} drivers are i.i.d.\ or mean-reverting: a trending or seasonal confounder --- the classic seasonality-confounded-treatment failure mode --- is therefore outside the released prior, and is the concrete content of the structured-driver extension we flag as future work (\S\ref{sec:conclusion}). Ground-truth semantics also differ by suite: the continuous-time generator simulates both arms under one shared noise realisation, so its targets are counterfactuals in the strict sense, whereas the discrete-time generators draw the interventional arm with independent noise. The released targets of \suite{dot-Identifiability-v1}, \suite{dot-RegimeSwitch-v1}, and \suite{dot-Generic-100k} are exact \emph{interventional} samples from the same SCM, not counterfactuals of the specific observational run. The prediction task is unchanged (predict the released target from the observational context and the intervention), but the discrete targets carry an independent noise realisation, which raises the irreducible-error floor and is consistent with the near-saturated squared-error metrics of \S\ref{sec:results}. Shared-noise discrete counterfactuals require only reusing the pre-sampled noise across arms and are planned for a versioned v2 release (v1 suites stay frozen). The genuine modelling assumptions are additive Markovian noise and no explicit measurement model (the continuous-time suite does support irregular jittered/Poisson observation grids, \S\ref{sec:scope}). Two further limitations of the headline experiment: the interventional/observational gap of \S\ref{sec:gap} is measured on held-out \emph{episodes} of the training prior (not under generator shift), and its primary study uses models trained \emph{without lagged edges} (a single-seed lagged replication preserves the gap, \S\ref{sec:gap}). Training-seed variance of the \code{OSC} gap is quantified over three seeds at $T{=}200$ (pooled seed std $0.005$, \S\ref{sec:gap}); the $T$-sweep and the \code{BTM} arms remain single-seed, and \S\ref{sec:repr} shows this model family can be seed-unstable in other settings. Finally, DoTime cannot yet validate real-world transportability beyond the proof-of-concept transfer of \S\ref{sec:repr}.
\paragraph{Responsible release.} Apache-2.0 (code) and CC-BY-4.0 (data); purely synthetic with no PII; fully reproducible from seeds; a frozen Zenodo DOI per suite; Croissant metadata; and a reported energy footprint for generating the $100$k suite (Appendix~\ref{app:datasheet}).

\section{Conclusion}\label{sec:conclusion}

DoTime is, to our knowledge, the first open, scalable, and theoretically grounded benchmark generator for multivariate time-series causal inference under interventions, spanning three intervention types with continuous-time windows, five counterfactual modes, eight identification structures, and four frozen suites with reference baselines.
Our headline experiment shows, across three training seeds, four trajectory lengths, and three identification structures, that an interventional foundation model outperforms an observational one of equal capacity on direction accuracy --- a gap the benchmark detects consistently (pooled $+0.09$, seed std $0.005$).
Future work includes prior-diversity extensions (broader mechanism families, structured exogenous drivers, prior-side episode filtering), partial observability, learned mechanism priors, and connections to causal world-model literature.

\appendix

\newpage

\section{Algorithms}\label{app:algo}
Algorithm~\ref{alg:sample} samples a temporal SCM and a paired observational-interventional trajectory; Algorithm~\ref{alg:sample} covers both rungs of ground truth (\S\ref{sec:framework}): the interventional arm either draws fresh noise (Level~2, an interventional twin) or reuses the observational arm's noise (Level~3, a counterfactual). As released, the continuous-time generator (\suite{dot-Continuous-v1}) takes the shared-noise branch; the discrete-time suites take the independent-noise branch (\S\ref{sec:limitations}). Algorithm~\ref{alg:cf} then reads the counterfactual target off the shared-noise run.

\begin{algorithm}[h]
\caption{Sample a paired (obs, int) trajectory}\label{alg:sample}
\begin{algorithmic}[1]
\STATE sample $G_0$ (upper-triangular mask over random topo order), $G_{1:K}$ (geometric decay)
\STATE sample mechanisms $f_i$ and noise scales; optionally rescale lag matrices to spectral radius $\le \lambda_{\max}$ (hardening config)
\STATE draw noise $\varepsilon_{1:T}$; simulate $X^{\text{obs}}$ via Eq.~\eqref{eq:recursion} with burn-in
\STATE sample intervention $(\text{target}, \text{window}, \text{value})$ under the chosen mode + positivity guard
\STATE \textbf{if} counterfactual (Level~3): simulate $X^{\text{int}}$ \emph{reusing} $\varepsilon_{1:T}$; \textbf{else} interventional (Level~2): draw fresh $\varepsilon'_{1:T}\sim P(\varepsilon)$ --- overriding the target over the window either way
\STATE \textbf{return} $(X^{\text{obs}}, X^{\text{int}}, \text{spec})$
\end{algorithmic}
\end{algorithm}

\begin{algorithm}[h]
\caption{Counterfactual target at a query}\label{alg:cf}
\begin{algorithmic}[1]
\STATE given the query (variable index for $Y$, time $t_q$) and the shared-noise (Level~3) interventional run
\STATE \textbf{return} $X^{\text{int}}_{t_q, Y}$ \COMMENT{exact counterfactual, since $\varepsilon$ is shared with the factual run}
\end{algorithmic}
\end{algorithm}

\section{Collapsed (continuous-time) structure views}\label{app:collapsed}
\begin{figure*}[t]
\centering
\begingroup
\tikzset{
  ct/.style   = {circle, draw=blue!55!black, fill=blue!15, minimum size=6.4mm,
                 inner sep=0pt, font=\small},
  cint/.style = {circle, draw=red!70!black, thick, fill=orange!75, minimum size=6.4mm,
                 inner sep=0pt, font=\small},
  cout/.style = {circle, draw=orange!85!black, thick, fill=white, minimum size=6.4mm,
                 inner sep=0pt, font=\small},
  chid/.style = {circle, draw=black!70, dashed, fill=white, minimum size=6.4mm,
                 inner sep=0pt, font=\small},
  ar/.style   = {-{Stealth[length=1.9mm]}, blue!55!black, semithick},
  cut/.style  = {-{Stealth[length=1.9mm]}, red!75!black, thick, dashed},
  slL/.style  = {ar,  out=160, in=200, looseness=7},   
  slR/.style  = {ar,  out=-20, in=20,  looseness=7},   
  slU/.style  = {ar,  out=70,  in=110, looseness=7},   
  slD/.style  = {ar,  out=250, in=290, looseness=7},   
  slLcut/.style={cut, out=160, in=200, looseness=7},   
  ttl/.style  = {font=\small\bfseries},
  fam/.style  = {font=\footnotesize\itshape, text=black!70},
}
\newcommand{\cpanel}[3]{%
  \begin{tikzpicture}[baseline=(current bounding box.north), x=1.35cm, y=1.15cm]
    \node[ttl] at (0.5,2.05) {#1}; \node[fam] at (0.5,1.68) {#2}; #3
  \end{tikzpicture}}
\setlength{\tabcolsep}{9pt}
\begin{tabular}{cccc}
\cpanel{Bi-Variate}{trivial (RCT)}{%
  \node[cint] (a) at (0,0) {$A$}; \node[cout] (y) at (1,0) {$Y$};
  \draw[ar] (a) -- (y); \draw[slLcut] (a) to (a); \draw[slR] (y) to (y);
}
&
\cpanel{Observed-Confounder}{back-door (null)}{%
  \node[ct] (x) at (0.5,0.6) {$X$}; \node[cint] (a) at (0,-0.4) {$A$};
  \node[cout] (y) at (1,-0.4) {$Y$};
  \draw[cut] (x) -- (a); \draw[ar] (x) -- (y);
  \draw[slLcut] (a) to (a); \draw[slU] (x) to (x); \draw[slR] (y) to (y);
}
&
\cpanel{Back-Door}{back-door}{%
  \node[ct] (x) at (0.5,0.6) {$X$}; \node[cint] (a) at (0,-0.4) {$A$};
  \node[cout] (y) at (1,-0.4) {$Y$};
  \draw[cut] (x) -- (a); \draw[ar] (x) -- (y); \draw[ar] (a) -- (y);
  \draw[slLcut] (a) to (a); \draw[slU] (x) to (x); \draw[slR] (y) to (y);
}
&
\cpanel{Confounder-Mediator}{back-door}{%
  \node[ct] (x) at (0.5,0.8) {$X$}; \node[cint] (a) at (-0.2,-0.2) {$A$};
  \node[ct] (m) at (0.5,-0.2) {$M$}; \node[cout] (y) at (1.2,-0.2) {$Y$};
  \draw[cut] (x) -- (a); \draw[ar] (a) -- (m); \draw[ar] (m) -- (y);
  \draw[ar] (x) to[bend left=20] (y);
  \draw[slLcut] (a) to (a); \draw[slU] (x) to (x); \draw[slD] (m) to (m); \draw[slR] (y) to (y);
}
\\[2.2em]
\cpanel{Mediator}{front-door}{%
  \node[cint] (a) at (0,0) {$A$}; \node[ct] (m) at (0.7,0) {$M$};
  \node[cout] (y) at (1.4,0) {$Y$};
  \draw[ar] (a) -- (m); \draw[ar] (m) -- (y);
  \draw[slLcut] (a) to (a); \draw[slD] (m) to (m); \draw[slR] (y) to (y);
}
&
\cpanel{Front-Door}{front-door}{%
  \node[chid] (u) at (0.5,0.8) {$U$}; \node[cint] (a) at (-0.2,-0.2) {$A$};
  \node[ct] (m) at (0.5,-0.2) {$M$}; \node[cout] (y) at (1.2,-0.2) {$Y$};
  \draw[cut] (u) -- (a); \draw[ar] (u) to[bend left=20] (y);
  \draw[ar] (a) -- (m); \draw[ar] (m) -- (y);
  \draw[slLcut] (a) to (a); \draw[slU] (u) to (u); \draw[slD] (m) to (m); \draw[slR] (y) to (y);
}
&
\cpanel{Instrumental-Variable}{IV}{%
  \node[chid] (u) at (0.6,0.8) {$U$}; \node[ct] (x) at (-0.4,-0.2) {$X$};
  \node[cint] (a) at (0.5,-0.2) {$A$}; \node[cout] (y) at (1.4,-0.2) {$Y$};
  \draw[cut] (x) -- (a); \draw[ar] (a) -- (y);
  \draw[cut] (u) -- (a); \draw[ar] (u) to[bend left=20] (y);
  \draw[slLcut] (a) to (a); \draw[slU] (u) to (u); \draw[slU] (x) to (x); \draw[slR] (y) to (y);
}
&
\cpanel{Unobserved-Confounder}{non-identifiable}{%
  \node[chid] (u) at (0.5,0.6) {$U$}; \node[cint] (a) at (0,-0.4) {$A$};
  \node[cout] (y) at (1,-0.4) {$Y$};
  \draw[cut] (u) -- (a); \draw[ar] (u) -- (y);
  \draw[slLcut] (a) to (a); \draw[slU] (u) to (u); \draw[slR] (y) to (y);
}
\end{tabular}
\endgroup

\caption{Collapsed continuous-time view of the same eight structures. Each node
is one variable; the self-loop is the OU drift's own-value term
($-\theta_v X_v$), and $u\!\to\!v$ means $u$ enters $v$'s drift. Red dashed
edges are severed by $\mathrm{do}(a)$ --- including $a$'s own self-loop, since a
hard intervention overrides the treatment's own dynamics for the whole window.
Node roles match Figure~\ref{fig:structures}: orange $=$ intervened, open $=$
queried outcome, dashed $=$ hidden.}
\label{fig:collapsed}
\end{figure*}
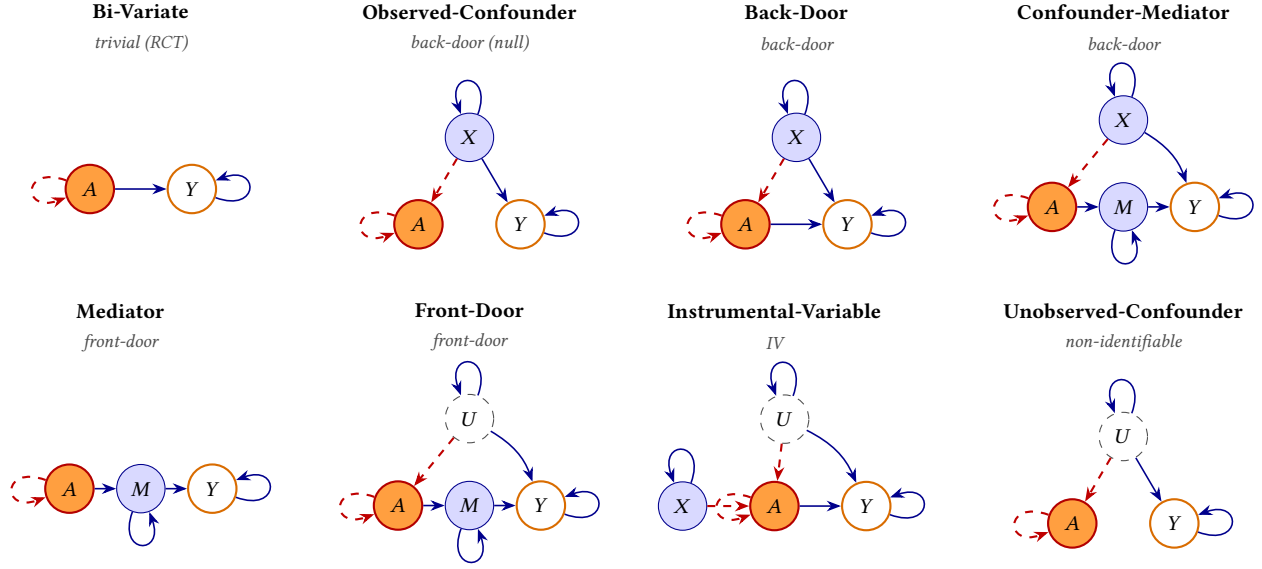

Figure~\ref{fig:structures} draws the eight identification structures
\emph{unrolled} over discrete slices, which is the right picture for the
discrete-time suites but redundant once the dynamics are continuous. Figure~\ref{fig:collapsed}
gives the complementary \emph{collapsed} view used for
\suite{dot-Continuous-v1} (cf.\ the causal graphs for systems of SDEs
of~\citet{boeken2026sde}): one node per variable, a self-loop standing for
the SDE's own-value dependence $-\theta_v X_v$, and an edge $u\!\to\!v$ wherever
$u$ enters $v$'s drift. The two views encode the same identification content;
the collapsed one makes clear that $\mathrm{do}(a)$ severs \emph{all} of $a$'s
incoming edges including its self-loop --- the continuous-time analogue of
holding the treatment fixed across the intervention window rather than at a
single step.

\section{Convergence theorem (restated)}\label{app:theory}
Under the six assumptions of \citet{thumm2026causal} (acyclic $G_0$; bounded, Lipschitz mechanisms; sub-Gaussian noise; spectral radius $<1$ after clipping;
positivity over the intervention support; and Markov order $K$), the simulated process is stationary after burn-in and the empirical counterfactual estimator is consistent. The full proof is reproduced from \citet{thumm2026causal}.

\paragraph{Scope on the released suites.} The result is stated conditionally, and
two of its assumptions are deliberately relaxed by the released v1.0.0 suites, so
we make the scope explicit rather than let the theorem be read as a guarantee
about the shipped data. First, the spectral-radius condition is enforced only by
the optional hardening configuration used for PFN training (\S\ref{sec:intro});
the released suites apply no such constraint. Computing the reduced-form
companion spectral radius $\rho$ directly from the sampled weight matrices of
$2{,}000$ freshly drawn generic SCMs (no simulation required), $\rho \ge 1$ in
$63\%$ of the $1{,}715$ draws for which a single companion matrix is defined.
This is precisely what the divergence disclosed in \S\ref{sec:validation} is:
$95\%$ of zeroed episodes have $\rho \ge 1$, and divergence is $48\%$ likely
given $\rho \ge 1$ against $4.6\%$ given $\rho < 1$. Because each mechanism
applies a nonlinearity to its linear pre-activation, this $\rho$ is exact only
for linear activations and an upper bound otherwise. Re-running the \emph{same}
$1{,}715$ sampled weight matrices with every mechanism restricted to a linear
activation --- which leaves $\rho$ unchanged and varies only the nonlinearity ---
makes the criterion exact, and the relationship becomes essentially
deterministic: divergence follows $\rho \ge 1$ in $96\%$ of cases, occurs in
\emph{none} of the $636$ draws with $\rho < 1$, and every zeroed episode has
$\rho \ge 1$. The difference between the two conditions is the prior's
nonlinearities doing stability work: being contractive, they rescue roughly half
of the $\rho \ge 1$ draws, which is why overall divergence falls from $52\%$ to
$27\%$ once they are enabled. Second,
\suite{dot-RegimeSwitch-v1} switches SCM parameters mid-trajectory by
construction --- it is non-stationary by design and admits no single companion
matrix (such episodes are the excluded remainder above), so the theorem does not
cover it and is not intended to. Conditional on an episode surviving divergence
detection, however, burn-in does what it is meant to: across the same sample the
per-variable variance ratio between the first and last $50$ retained steps has
median $1.01$ with $94\%$ of variables within a factor of two, and the
standardized mean drift $|\bar{x}_{\text{late}} - \bar{x}_{\text{early}}|/
\hat\sigma_{\text{early}}$ has median $0.14$ with $96\%$ below $0.5$. The
theorem therefore describes the hardened, non-switching configuration; the
released suites trade that guarantee for prior diversity, and users needing the
stationary regime can rebuild with \code{--stability-retries}
(\S\ref{sec:validation}). Both measurements reproduce from the shipped
\code{dotime-diagnose-stationarity} console script; the full output is released as \\ \code{results/reference/stationarity\_diagnostic.json} (and \code{\_identity.json} for the linear-activation condition).

\section{Hyperparameters and configuration}\label{app:hparams}
Default prior: $N\!\sim\!\mathcal{U}\{3,\dots,10\}$, $K\!\sim\!\mathcal{U}\{1,\dots,3\}$, edge
probability $\mathrm{Beta}(2,5)$, geometric lag decay $\gamma$ (default $0.7$), dropout (hidden)
probability up to $0.3$, burn-in $50$. Per-suite seeds and the  regime / continuous
configurations are recorded in \code{scripts/release\_config.yaml}; the build
manifest records the config hash and package version. The counterfactual-mode OOD
ratios and mean effect magnitudes are tabulated per mode in the released metadata.

\section{Per-structure breakdown}\label{app:perstruct}

\begin{figure*}[t]
\centering
\includegraphics[width=0.94\textwidth]{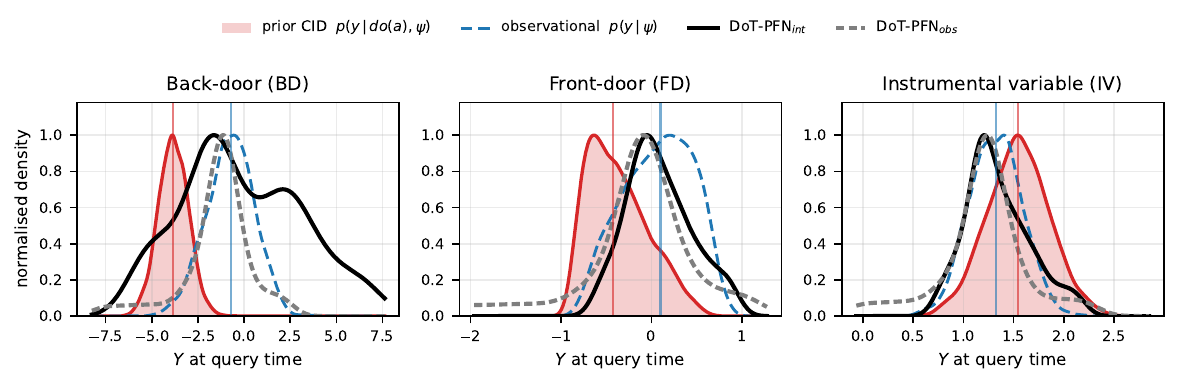}
\caption{Conditional interventional distribution (CID) at the query time for one
representative TSCM per structure (\code{OSC} prior; the SCM seed is chosen by a
deterministic scan targeting a standardized do-effect of $2.5\sigma$ --- realized
$2.5\sigma$ / $1.3\sigma$ / $0.9\sigma$ for BD/FD/IV). Red, filled: the
ground-truth CID $p(y \mid \mathrm{do}(a), \psi)$ from $3{,}000$
independent-noise interventional rollouts of the same SCM (Level-2 semantics,
\S\ref{sec:framework}); blue dashed: the observational density
$p(y \mid \psi)$; vertical lines mark the two medians. Black:
\code{DoT-PFN}$_{\mathrm{int}}$'s predictive density; grey dashed:
\code{DoT-PFN}$_{\mathrm{obs}}$ at identical capacity, both conditioned on the
same representative observational history. The interventional model shifts its
mass toward the CID while its observational twin stays on the observational
density, and the visual separation is largest for back-door and vanishes for IV
--- the same ordering as the measured gaps in Table~\ref{tab:gap}. This is a
qualitative single-SCM illustration, not an aggregate: neither model recovers
the full density.}
\label{fig:cid}
\end{figure*}

Table~\ref{tab:gap} reports the \code{OSC} prior. On the \code{BTM} prior the same
$T$-sweep gives gaps of $+0.05$ to $+0.07$ (back-door), $+0.03$ to $+0.09$
(front-door), and $-0.02$ to $+0.01$ (instrumental-variable), pooling to $+0.02$
to $+0.05$ across $T \in \{200, 500, 1000, 2000\}$ (per-cell standard errors
$\approx 0.03$, Table~\ref{tab:gapbtm}) --- the same non-negative trend as \code{OSC} at
smaller magnitude, consistent with the \code{BTM} generator producing weaker, noisier
interventional signals; the IV gap is not distinguishable from zero. Full
per-structure RMSE, NMSE, MAE, direction accuracy, and standard errors ship in the
released results JSONs.

\begin{table}[t]
\caption{\code{BTM} prior: direction-accuracy gap (int $-$ obs, $\pm$ standard
error) of the structure-matched Do-Over-Time-PFN across trajectory lengths (same
protocol as Table~\ref{tab:gap}: held-out episodes, no lagged edges; single
training run per arm --- \code{BTM} seed variance not separately quantified). \tgapnote\ \gapnote}
\label{tab:gapbtm}
\small
\setlength{\tabcolsep}{3pt}
\begin{tabular}{lcccc}
\toprule
Structure & $T{=}200$ & $T{=}500$ & $T{=}1000$ & $T{=}2000$ \\
\midrule
back\_door             & $+0.07\sd{.03}$ & $+0.05\sd{.03}$ & $+0.05\sd{.03}$ & $+0.05\sd{.03}$ \\
front\_door            & $+0.05\sd{.03}$ & $+0.09\sd{.03}$ & $+0.06\sd{.03}$ & $+0.03\sd{.03}$ \\
instrumental\_variable & $-0.01\sd{.03}$ & $+0.01\sd{.03}$ & $-0.02\sd{.03}$ & $-0.01\sd{.03}$ \\
\midrule
\textbf{pooled}        & $\mathbf{+0.04}$ & $\mathbf{+0.05}$ & $\mathbf{+0.03}$ & $\mathbf{+0.02}$ \\
\bottomrule
\end{tabular}
\end{table}

\paragraph{Exploratory: does the gap track identifiability?}
A natural question is whether the interventional advantage vanishes where the
effect is not identifiable. As a single-seed probe we trained the same
structure-matched pair on two further structures and evaluated at $T{=}200$:
\code{observed\_confounder} (identified, back-door family) gives a gap of
$+0.03\sd{.03}$ ($0.92$ vs.\ $0.89$) and \code{unobserved\_confounder}
(non-identifiable by design) gives $+0.03\sd{.02}$ ($0.89$ vs.\ $0.85$) ---
small, similar, and \emph{not} vanishing under non-identifiability. Two caveats
frame this. First, both structures are near-saturated: even the observational
model signs the effect $\ge\!0.85$ of the time, leaving little headroom for any
gap --- consistent with the pattern that the gap is largest where the task is
hardest (back-door at $0.62$--$0.78$). Second, direction accuracy scores the
\emph{sign} of the effect, and sign recovery is a strictly weaker task than
effect identification: an unobserved confounder biases the magnitude of the
naive estimand but need not flip its sign under this generator's effect-size
distribution. A magnitude-sensitive metric may separate the identifiable and
non-identifiable structures more sharply; we release the per-cell JSONs
(\code{s9ho\_extra/}) and leave that refinement to future work.

\section{Datasheet (abbreviated)}\label{app:datasheet}
Following \citet{gebru2021datasheets}: \emph{Motivation} --- to benchmark
interventional/counterfactual time-series estimation. \emph{Composition} ---
purely synthetic temporal SCM trajectories with exact ground truth; no human
subjects, no PII. \emph{Collection} --- generated from a committed seed-fixed
script. \emph{Uses} --- training and evaluating causal foundation models; not for
clinical decision-making. \emph{Distribution} --- Zenodo (DOI) + Hugging Face,
under CC-BY-4.0. \emph{Maintenance} --- versioned suites; the GitHub repo tracks
issues and the leaderboard. \emph{Energy footprint} --- generation is CPU-only (no
GPU). A measured $10$k-episode slice of \suite{dot-Generic-100k} built in $157$\,s
wall-clock ($15$ workers, ${\approx}14$ active cores) on a $16$-thread AMD Ryzen~7
7840HS laptop; the full $100$k suite therefore takes ${\approx}26$\,min of
wall-clock time (${\approx}6.2$ core-hours of CPU work). Energy is
package-power $\times$ wall-clock --- the ${\sim}45$\,W figure is the whole-CPU
package draw, already covering all active cores, so it is \emph{not} multiplied by
core-hours: $45$\,W $\times\ 0.44$\,h ${\approx}0.02$\,kWh (computed from rated
package power, as direct RAPL (per-package energy) counters were unavailable).

\section{Croissant metadata (sample)}\label{app:croissant}
Each suite ships a \code{croissant.json} JSON-LD descriptor with the file objects
(parquet shards with md5), the record-set fields (\code{x\_obs}, \code{x\_int},
\code{y\_true}, \code{structure}, \code{tier}), license, and version, conforming
to the Croissant specification~\cite{croissant2024}.

\section{Reproducibility checklist}\label{app:repro}
Code and seeds are released; \code{pip install dotime} installs the
generator and loaders on a clean Python 3.10--3.12 environment; all four suites
round-trip through \code{load\_benchmark} with checksum verification; the
reference baselines and the Table~\ref{tab:results}/Table~\ref{tab:fmgap} rows
reproduce from the evaluation harness shipped inside the package
(\code{dotime-eval-reference}, \code{dotime-eval-pfn},
\code{dotime-eval-tabpfn}, \code{dotime-eval-chronos}); the Appendix~\ref{app:theory}
stationarity diagnostics reproduce from \code{dotime-diagnose-stationarity}; and
the Do-Over-Time-PFN checkpoints are released on Hugging Face
(\code{OSC}~$=$~\code{s9ho}, \code{BTM}~$=$~\code{s9btm}), including the general
\code{s9ho\_all\_*} pair behind Table~\ref{tab:results}/Table~\ref{tab:fmgap}, the
per-structure \code{s9ho\_*}/\code{s9btm\_*} pairs behind
Tables~\ref{tab:gap}/\ref{tab:gapbtm}, the seed-43/44 checkpoints behind the
three-seed result (\S\ref{sec:gap}), the lagged-replication pairs
(\code{s9ho\_*\_lag}), and the two extra-structure pairs behind
Appendix~\ref{app:perstruct}. The training codebase, the structure-matched
evaluation drivers (\code{scripts/analyze\_s9ho.py}), and the
\code{OSC}/\code{BTM} hardening configurations are released (anonymized) at
\url{https://anonymous.4open.science/r/do-over-time-pfn-EF8D}, and every reported
cell's output metrics ship as JSONs under
\code{results/reference/structure\_matched/}.

\begin{acks}
We utilized Claude Code to assist in writing, scripting, and generating the PyPI package. All AI-generated code was manually audited, tested and validated by the authors.
\end{acks}

\bibliographystyle{ACM-Reference-Format}
\bibliography{references}

\end{document}